\documentclass[preprint,12pt]{elsarticle}

\usepackage{amssymb}
\usepackage{algorithm}
\usepackage{algorithmic}
\usepackage{amsmath}
\usepackage{theorem}
\usepackage{amssymb}
\usepackage{booktabs}
\usepackage{graphicx}
\newtheorem{theorem}{Theorem}[section]
\newtheorem{mytheorem}{Theorem}
\newtheorem{myproof}{Proof}

\usepackage{multirow}
\usepackage{color}

\journal{NeuralComputing}

\begin{document}

\begin{frontmatter}

\title{Deep Multi-Threshold Spiking-UNet for Image Processing}

\author[1]{Hebei Li}
\author[1]{Yueyi Zhang\corref{mycorrespondingauthor}}
\cortext[mycorrespondingauthor]{Corresponding author}
\ead{zhyuey@ustc.edu.cn}
\author[1]{Zhiwei Xiong}
\author[1,2]{Xiaoyan Sun}

\affiliation[1]{organization={MoE Key Laboratory of Brain-inspired Intelligent Perception and Cognition, University of Science and Technology of China},
            city={Hefei},
            citysep={},
            postcode={230026}, 
            country={China}}

\affiliation[2]{organization={Institute of Artificial Intelligence, Hefei Comprehensive National Science Center},
            city={Hefei},
            citysep={},
            postcode={230088}, 
            country={China}}

\begin{abstract}
    U-Net, known for its simple yet efficient architecture, is widely utilized for image processing tasks and is particularly suitable for deployment on neuromorphic chips. This paper introduces the novel concept of Spiking-UNet for image processing, which combines the power of Spiking Neural Networks (SNNs) with the U-Net architecture. To achieve an efficient Spiking-UNet, we face two primary challenges: ensuring high-fidelity information propagation through the network via spikes and formulating an effective training strategy. To address the issue of information loss, we introduce multi-threshold spiking neurons, which improve the efficiency of information transmission within the Spiking-UNet. For the training strategy, we adopt a conversion and fine-tuning pipeline that leverage pre-trained U-Net models. During the conversion process, significant variability in data distribution across different parts is observed when utilizing skip connections. Therefore, we propose a connection-wise normalization method to prevent inaccurate firing rates. Furthermore, we adopt a flow-based training method to fine-tune the converted models, reducing time steps while preserving performance. Experimental results show that, on image segmentation and denoising, our Spiking-UNet achieves comparable  performance to its non-spiking counterpart, surpassing existing SNN methods. Compared with the converted Spiking-UNet without fine-tuning, our Spiking-UNet reduces inference time by approximately 90\%. This research broadens the application scope of SNNs in image processing and is expected to inspire further exploration in the field of neuromorphic engineering. The code for our Spiking-UNet implementation is available at https://github.com/SNNresearch/Spiking-UNet.
\end{abstract}

\end{frontmatter}
\section{Introduction}
    U-Net \cite{ronneberger2015u} has emerged as a popular choice for a wide range of image-related applications \cite{xiao2020global, zhou2020hierarchical, liu2022video, nazir2021ecsu}, particularly on tasks involving pixel-wise prediction such as image segmentation and denoising. One of the key characteristics of U-Net is its unique use of skip connections, which perform concatenation operations between the corresponding layers of the encoder and the decoder. This strategy allows it to preserve full resolution and maintain contextual semantic information during the feature learning process \cite{cciccek20163d}, and to retain the fine-grained details of the target objects. These advantages help U-Net achieve promising performance particularly when limited training samples are available. However, the pixel-wise applications of U-Net on the rapidly emerging neuromorphic devices remain limited.

    In recent years, Spiking Neural Networks (SNNs) \cite{sengupta2019going, liao2023convolutional, zhang2019tdsnn, yang2022training} have emerged as a rapidly growing field of research, driven by their unique ability to emulate the efficient and intricate working scheme of biological nervous systems. In contrast to traditional artificial neural networks, SNNs utilize spiking neurons which convey information through asynchronous, discrete spikes that closely resemble the behavior of biological neurons. This spiking mechanism endows SNNs with inherent advantages such as low energy consumption, fast inference, and event-driven processing \cite{akopyan2015truenorth, davies2018loihi, pei2019towards, furber2014spinnaker}. Owing to these benefits, SNNs have attracted substantial interest for their potential in energy-constrained applications such as the Internet of Things (IoT), small satellites, and robotics. By leveraging these bio-inspired spiking mechanisms, SNNs open promising avenues for efficient and high-performance computation across various domains.

    Despite the promising advantages of SNNs, implementing them effectively in tasks requiring real-time or rapid inference presents significant challenges. The first of these challenges is the need to maintain efficient information propagation within the network using spikes. To tackle this, researchers have explored the use of ResNet-based deep network topologies \cite{fang2021deep} and the incorporation of spiking neurons \cite{feng2022multi, guo2022real,fang2021incorporating}. These advancements have led to the development of deep SNNs, which have shown promising results. The second challenge pertains to the design of effective training strategies. Two primary approaches emerge: converting ANNs to SNNs and training SNNs directly. While these methods have primarily been applied to classification tasks \cite{wu2023dynamic, xu2023ultra, chen2022adaptive}, the exploration of SNNs in deep U-Net architectures for pixel-wise tasks remains an under-explored area. 
    
    In this paper, we propose the Spiking-UNet, an efficient integration of SNNs and the U-Net architecture for pixel-wise tasks, specifically image segmentation and denoising. To address the challenge of information propagation using spikes, we introduce multi-threshold spiking neurons that fire spikes at different thresholds, enhancing performance in a short time window. This mechanism promotes accurate spike propagation to subsequent layers, ensuring effective information flow. For effective training Spiking-UNet, we construct our model by converting a pre-trained U-Net model and subsequently fine-tuning it. During the ANN-SNN conversion, we observe that the data distribution from different parts of skip connections has significant variability, leading to inconsistent firing rates in Spiking-UNet. To overcome this, we propose a connection-wise normalization strategy, which equalizes the firing rates across skip connections, thereby ensuring more consistent and effective information transmission. In terms of fine-tuning, the traditional Back Propagation Through Time (BPTT) approach commonly used in training SNNs is computationally demanding. To mitigate this, we revise a training method that utilizes an accumulated spiking flow approach to more efficiently update the weights of the converted Spiking-UNet. To validate the effectiveness of our proposed Spiking-UNet, we conduct image segmentation experiments on the DRIVE \cite{staal2004ridge}, EM segmentation \cite{cardona2010integrated}, and CamSeq01 datasets \cite{fauqueur2007assisted}, as well as image denoising on the BSD68, and CBSD68 datasets \cite{martin2001database}. Experimental results demonstrate that our Spiking-UNet not only exceeds the existing SNN methods but achieves comparable performance to the corresponding U-Net.
    
    In summary, the key contributions of our work can be outlined as follows:

    \begin{itemize}
        \item We introduce the Spiking-UNet, a deep SNN utilizing U-Net architecture, trained via ANN-SNN conversion and fine-tuning, for image processing.
        \item To improve the accuracy of information representation, we introduce the multi-threshold spiking neuron and offer a criterion for establishing the optimal threshold values. This ensures maximum efficiency in capturing spiking activity.
        \item We propose a connection-wise normalization strategy to address the significant discrepancies in activation distribution introduced by skip connections. This approach aligns the spiking rates with the features of the original U-Net model, thus ensuring faithful information representation in Spiking-UNet.
        \item We extensively evaluate the performance of Spiking-UNet on image segmentation and denoising tasks using multiple datasets. Our experiments demonstrate that Spiking-UNet not only outperforms existing SNN methods but also achieves performance comparable to the traditional U-Net model .
    \end{itemize}

\section{Related Work}
  \subsection{Spiking Neuron Model}
    The majority of spiking neural networks predominantly employ the Integrate-and-Fire (IF) model \cite{abbott1999lapicque} or the Leaky-Integrate-and-Fire (LIF) model \cite{doutsi2021dynamic}. However, these spiking neuron models have shown limitations in efficient information transmission, as highlighted in recent studies \cite{li2022brain, wu2021liaf}. To improve this, Guo {et al.} \cite{guo2022real} employed real-valued spikes (RS) neurons for training and binary spikes for inference. This technique enhances SNNs' information representation capabilities and demonstrated superior performance on various datasets by the re-parameterization technique. In addition, Lang {et al.} \cite{feng2022multi} proposed a multi-level spiking neuron (ML) for improving the efficiency of spiking neural networks. This method enhances gradient propagation and the expression ability of neurons in these networks. As part of our methodology, we propose the multi-threshold spiking neuron, which fires multiple spikes upon reaching different thresholds. This approach aligns the output of the spiking neuron with the activation of the artificial neural network during the conversion process, thereby streamlining the training after conversion.
    
  \subsection{Deep SNN For Image Processing}
    There are two primary strategies for constructing efficient deep SNNs: one involves converting ANNs to SNNs, and the other entails direct SNN training. In the conversion process, pre-trained ANNs are transformed into SNN counterparts. Diehl \emph{et al.} \cite{diehl2015fast} proposed a layer-wise weight normalization method combined with threshold balancing to regulate firing rates on image classification task. Likewise, Rueckauer \emph{et al.} \cite{rueckauer2017conversion} provided some spiking equivalents of common ANN operations and introduced a strategy for selecting robust scale factors for layer-wise normalization. Yan \emph{et al.} \cite{yan2021near} developed a clamped and quantized training method to minimize loss during the conversion process in image classification tasks. Moreover, a recent study in object detection introduced signed neurons with imbalanced thresholds and successfully converted a Tiny YOLO model to an SNN version \cite{kim2018spiking}. 

    Directly training SNNs can be conducted via supervised or unsupervised methods. The classical unsupervised training method for SNNs is Spike-Timing-Dependent-Plasticity (STDP). Diehl \emph{et al.} \cite{diehl2015unsupervised} applied STDP to train an image classification network. Liu \emph{et al.} \cite{liu2020unsupervised} used the STDP method to train an SNN for object recognition with event representation. Liu \emph{et al.} \cite{liu2019deep} integrated the STDP method with adaptive thresholding for video-based disguise face recognition. However, supervised methods confront the issue of non-differentiable spiking neuron transfer functions. To address this, Lee \emph{et al.} \cite{lee2016training} proposed a method to smooth spikes and calculate the gradient of each layer using chain rules. Wu \emph{et al.} \cite{wu2018spatio} presented a supervised method with spatial-temporal backpropagation (STBP) to train SNN models for image classification, leveraging the surrogate gradient. More recently, Chakraborty \emph{et al.} \cite{chakraborty2021fully} used a hybrid method combining STDP and STBP to address object detection. Nevertheless, these training methods based on BPTT consume significant memory due to the need to record each timestep's activation. Kim \emph{et al.} \cite{kim2022beyond}  explored the use of SNN for semantic segmentation, presenting the Spiking-FCN architecture. They showed that direct training Spiking-FCN with surrogate gradient learning yielded better performance and lower latency compared to traditional ANN-to-SNN conversion methods. Wu \emph{et al.} \cite{wu2021training} proposed an efficient spike integration method, ASF-BP, to calculate gradients in image classification, thus reducing resource consumption. In our work, we modify the ASF-BP method to train our SNN model, which substantially reduces the training time for our Spiking-UNet.
    
    \subsection{Combined SNN and U-Net Architecture}
    The U-Net architecture has found diverse applications within the SNN field. Zhu \emph{et al.} \cite{zhu2022event} proposed a potential-assisted spiking neural network combined with U-Net for event-based video reconstruction. Lee \emph{et al}. \cite{lee2020spike} presented a hybrid ANN-SNN U-Net architecture, designed to address the issue of spike vanishing. Hagenaars \emph{et al.} \cite{hagenaars2021self} introduced the self-supervised learning of hybrid ANN-SNN U-Net for event-based optical flow. Rançon \emph{et al.} \cite{ranccon2022stereospike} proposed the novel SNN readout for modified U-Net for depth estimation. Cuadrado \emph{et al.} \cite{cuadrado2023optical} combined the 3D encoder U-Net and SNN to extract the temporal context information for optical flow. Patel \emph{et al.} \cite{patel2021spiking} converted the shallow U-Net to the corresponding spiking version for image segmentation and deployed on a Loihi chip. However, this work only modified the ReLU activation during training U-Net and utilized LIF spiking neurons in the SNN. Compared with Patel's work, our deep Spiking-UNet utilizes multi-threshold spiking neurons and efficient training after conversion. Besides, our Spiking-UNet solves a potential problem of layer-wise normalization and evaluates across various datasets.

\section{Spiking-UNet}
    \begin{figure*}
    \centering
        \includegraphics[width=1.0\textwidth]{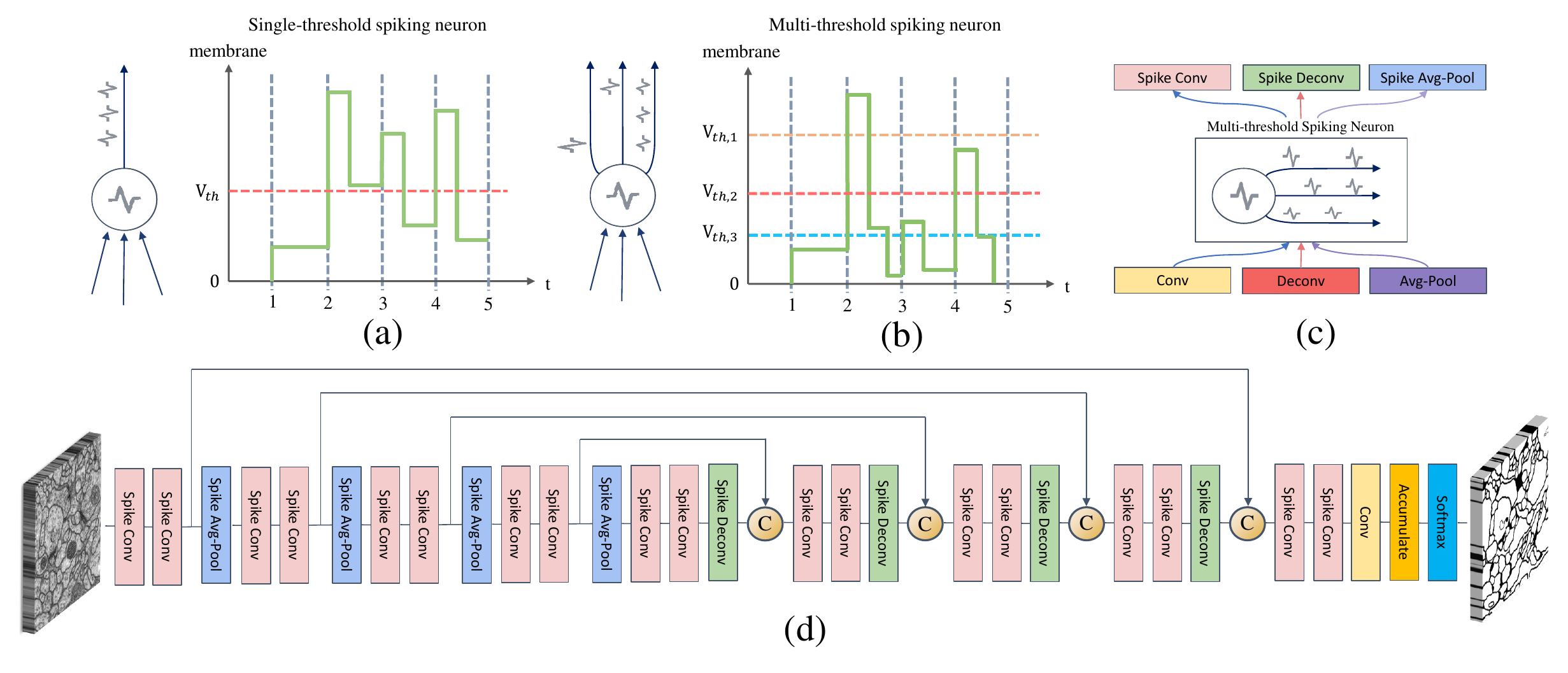}
        \caption{(a) A single-threshold spiking neuron with only one threshold, is poor in information representation. (b) A multi-threshold spiking neuron with multiple thresholds, has enhanced information transmission ability. (c) A multi-threshold spiking neuron in convolution in SNN. (d) The Spiking-UNet architecture for the image segmentation task, basically follows the classic U-Net architecture. The information flow in Spiking-UNet is discrete spikes instead of continuous values. Average pooling is utilized to replace max pooling for convenient conversion.}
        \label{fig:1}
    \end{figure*}
    \subsection{Overview}
      Following the ANN-SNN-Conversion principle, Spiking-UNet adopts the same architecture as the pre-trained U-Net model described in \cite{ronneberger2015u}. The configuration of the selected U-Net model closely aligns with the classic setting. The encoder component incorporates four down-sampling stages, while the decoder component utilizes four up-sampling stages. Notably, our baseline model differs from the classic U-Net by employing average pooling operations instead of max pooling operations \cite{rueckauer2017conversion} because average pooling is the form of convolution, which can simulate the relationship between the input and output of the data compared with max pooling. The architecture of the converted Spiking-UNet is depicted in Figure \ref{fig:1}(d).

      To construct our Spiking-UNet, we first train the U-Net model using conventional back-propagation with supervision labels. Once we have the pre-trained model, we replace the ReLU activation in the ANN with the multi-threshold spiking neuron model. We adjust the weights of kernels and biases in different convolutional layers using weight normalization methods, which will be described in the following subsections. Subsequently, we fine-tune the Spiking-UNet further to reduce the time window. For image segmentation tasks, the segmentation output is obtained by accumulating the spikes through the argmax function in the output layer. In the case of image denoising, the network output is the averaged membrane potential, treated as a noise map. In the first layer of the Spiking-UNet, we adopt static coding with the analog current as the input, following the approach presented in \cite{rueckauer2017conversion, kim2020spiking}. The analog current remains constant throughout the entire time window. In the subsequent layers, the inputs are spikes fired by neurons in the preceding layer at the previous time step.

    \subsection{Multi-Threshold Spiking Neuron}
     In the Spiking-UNet, we utilize the Integrate-and-Fire (IF) neuron model as the foundational neuron model for constructing the SNNs. The traditional IF neuron, which is a single-threshold spiking model, accumulates membrane potential when it receives input current. When the membrane potential exceeds the threshold voltage, the IF neuron emits a spike to its connected neurons in the next layer. After emitting the spike, the membrane potential of the neuron decreases, and the dynamics can be described by the following equations:
     \begin{align}
      H^l(t) &= V^l(t-1) + I(t), \\
      s^l(t) &= \Theta(H^l(t) - V_{th}), \\
      V^l(t) & = H^l(t) - V_{th}s^l(t), \\
      o^l(t) &= s^l(t),
    \label{Eq:1}
    \end{align}
    where $I^l(t)$, $\Theta(x)$, $s^l(t) \in \{0, 1\}$, and $o^l(t)$ represent the input current, Heaviside step function, spike of the neuron, and output of the spiking neurons in the $l^{th}$ layer at time $t$, respectively. $H^l(t)$, $V^l(t)$ represent the membrane potential after accumulation and after firing spikes in the $l^{th}$ layer at time $t$. $V_{th}$ denotes the threshold. However, this neuron model faces a challenge. If the threshold is set too low, input currents that surpass the threshold are disregarded, leading to significant conversion loss. On the other hand, if the threshold is set too high, small input currents require a longer time to trigger a spike.

     To address this issue, we introduce a multi-threshold spiking neuron that enables more precise representation in a short time window. The multi-threshold spiking neuron's membrane potential encounters a series of thresholds from high to low, such as (1.0, 0.8, 0.6). Once the membrane potential exceeds the first threshold, the neuron fires a spike for this threshold and the membrane potential is reduced by this threshold value. After that or if the threshold is not achieved, the membrane potential will face the next threshold, which is smaller than the previous threshold. This procedure repeats until all the thresholds are processed. This coarse-to-fine approach of the multi-threshold spiking neuron promotes information flow in deep SNNs. Mathematically, the multi-threshold spiking neuron can be described as follows:
     \begin{align}
      H^l(t) &= V^l(t-1) + I(t)\\
      s_i^l(t) &= \Theta(H^l(t) -\sum_{j=1}^{i-1} V_{th,j}s^l_j(t) - V_{th, i})\\
      V^l(t) &= H^l(t) - \sum_{i}V_{th,i}s_i^l(t) \\
      o^l(t) &= \sum_iV_{th,i}s_i^l(t)
      \label{Eq:2}
    \end{align}
    where $V_{th,i}$ is the $i^{th}$ threshold value of the spiking neuron, and $s_i^l(t)$ is the corresponding spike. The output $o^l(t)$ is the weighted sum of spikes. To better approximate the ANN activation, it is crucial to carefully choose threshold values. Based on the Theorem \ref{Theorem:1}, we can derive the relationship between the $V_{th, i+1}$ and $V_{th, i}$. The proof of Theorem \ref{Theorem:1} is provided in the Appendix section.
      
      \begin{theorem}
        Consider a multi-threshold neuron model with a membrane voltage $V$ that follows a uniform distribution in the range $[0, 1]$ ($V \sim U[0, 1]$). Let $N$ thresholds be denoted as $V_{th,1}, V_{th,2}, ..., V_{th,N}$, subject to the constraints $V_{th,1} > V_{th,2} > ... > V_{th,N-1} > V_{th,N}$ and $\sum_i V_{th,i} \leq 1$.

        Suppose the multi-threshold spiking neuron can fire only one spike for each threshold. Then the membrane voltage $V$ can be expressed as $V = \alpha_1 V_{th,1} + \alpha_2 V_{th,2} + ... + \alpha_N V_{th, N} + r(V)$, where $\alpha_i \in \{0, 1\}$ indicates whether a spike is fired at each threshold, and $r(V)$ represents the errors between the actual membrane voltage $V$ and the output weighted sum of spikes.

        The optimal series of thresholds that minimizes the error expectation satisfies $V_{th,i+1}=V_{th,i}/{2}$ and $V_{th, N} = 1/{2^N}$, yielding a minimal error expectation of $r^* = 1/{2^{N+1}}$.
        \label{Theorem:1}
      \end{theorem}
     
    \subsection{Weight Normalization}
      \begin{figure*}[t]
          \centering
          \includegraphics{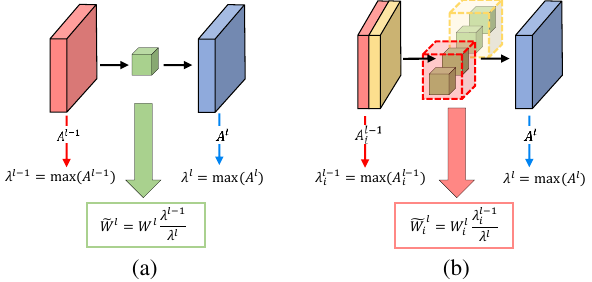}
          \caption{Illustration of different weight normalization methods. (a) Layer-wise Normalization (b) Connection-wise Normalization. Layer-wise normalization utilizes only one ratio of the maximum input activation to the maximum output activation, applying for all weights. Connection-wise normalization considers the normalization for the concatenation operation, which computes the ratio of the maximum activation of the input part to the maximum output activation, applying for the corresponding weights.}
          \label{fig:2}
      \end{figure*}
      To ensure accurate representation and transmission of input magnitude in converted SNNs, it is essential to ensure that spiking neurons emit spike trains without loss. Weight normalization is a key factor in scaling weights and biases to maintain appropriate spike rates. In this section, we introduce two normalization methods: layer-wise normalization and a novel connection-wise normalization. The connection-wise normalization method specifically addresses the issue of inconsistent spike rates introduced by skip connections in the U-Net architecture. Figure \ref{fig:2} visually illustrates these two normalization methods.

    \subsubsection{Layer-wise normalization}
      The layer-wise normalization technique, initially proposed in \cite{diehl2015fast} and extended in \cite{rueckauer2017conversion}, employs a scale factor to compute the normalized weights. This scale factor is determined by the ratio between the maximum activation of the previous layer and the maximum activation of the current layer. Mathematically, the normalized parameters are defined as follows:
      \begin{equation}
      \tilde{W}^l =W^l\frac{\lambda^{l-1}}{\lambda^{l}},\quad \tilde{b}^l = \frac{b^l}{\lambda^l}
      \end{equation}
      where $\tilde{W}^l$, $W^l$, $\lambda^l$, and $b^l$ denote the normalized weights, pre-trained weights, the maximum activations calculated from the training dataset in ANNs, and bias in the $l^{th}$ layer, respectively. To eliminate outliers, Rueckauer \emph{et al.} utilized the $99.9 \%$ percentile of the maximum activation $\lambda^l$ \cite{rueckauer2017conversion}, which was also adopted in Spiking-UNet.
    \subsubsection{Connection-wise Normalization}
      We observed there is a drawback in layer-wise normalization when it is applied to convert the skip connections of U-Net. This drawback arises from the significant variation in maximum activations across different parts of the concatenation. The use of a universal conversion scale factor leads to unbalanced weights for the skip connection part with smaller maximum activations. 
      
      To address this issue, we propose a connection-wise normalization method. Firstly, we calculate the scale factor between the maximum activations of different parts of the concatenation with skip connections and the maximum activation of the layer after the skip connections. Then, we divide the weights into different parts based on the concatenation with skip connections. We multiply the corresponding scale factor by the weights to obtain the normalized weights, formulated as:
      \begin{equation}
          \tilde{W}_{i}^l=W_{i}^l\frac{\lambda_i^{l-1}}{\lambda^l},\quad \tilde{b}^l=\frac{b^l}{\lambda^l}
      \end{equation}
      where $i$ represents the index of the part in the concatenation with skip connections. Our connection-wise normalization method specifically addresses the layer-wise problem associated with skip connections. 
        \begin{figure*}[t]
            \centering
            \includegraphics[width=1.0\textwidth]{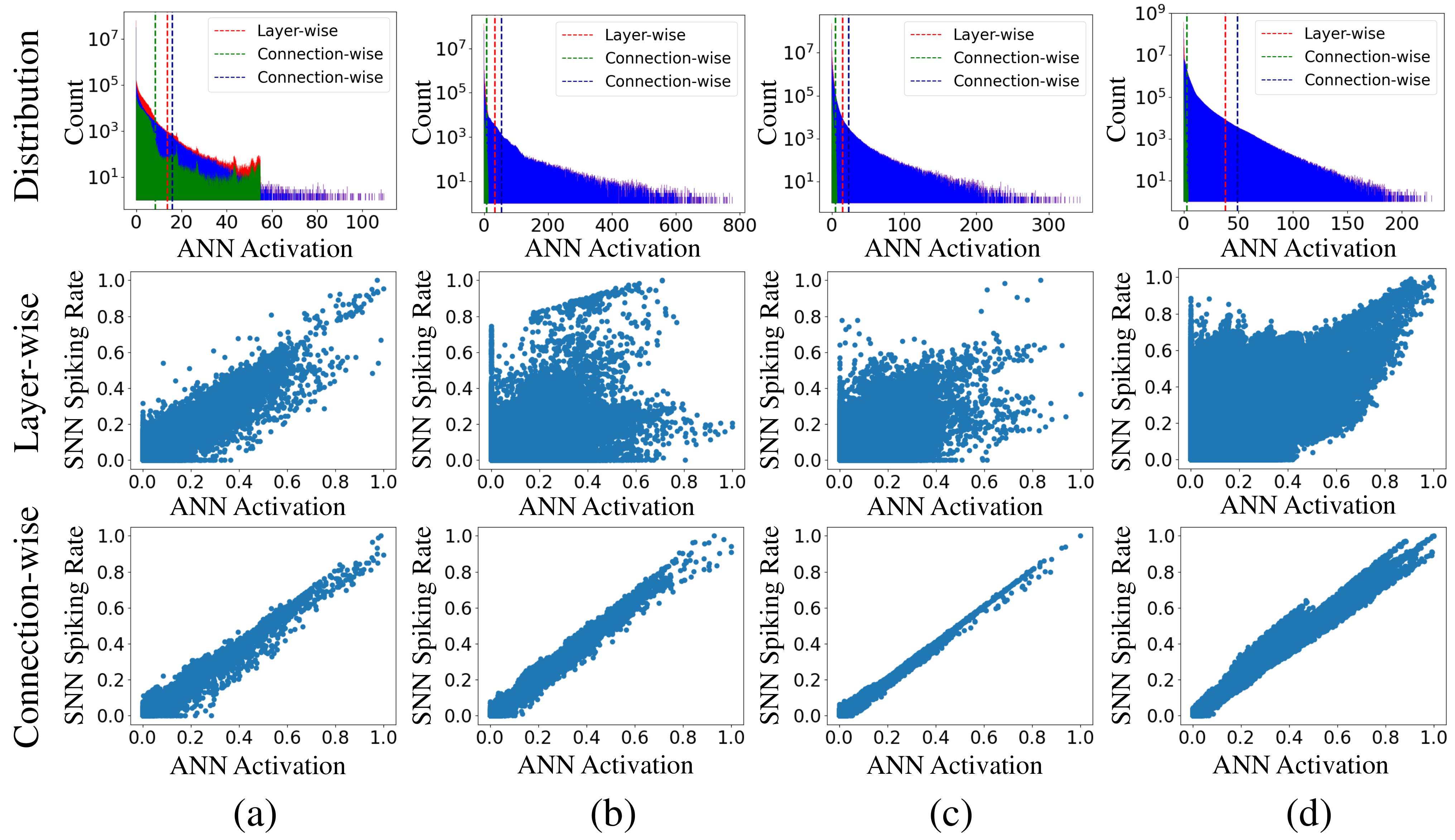}
            \caption{Illustration of activation distribution of U-Net and correlation between activation and spiking rate for layer-wise and connection-wise normalization. (a - d) are the 11th, 13th, 15th and 17th layers after skip connections, respectively.}
            \label{fig:3}
        \end{figure*}
      Figure \ref{fig:3} illustrates the activation distribution of U-Net and the relationship between activation and spiking rate. It can be observed that significant variations in activation distribution lead to variations in the maximum activation across different parts of the concatenation. The goal of ANN-SNN conversion is to establish a direct mapping between the spike rates in SNN and the activation values in ANN. Hence, an effective conversion process should ensure a linear relationship between the activation of ANN and the spike rate of SNN. However, layer-wise normalization disrupts this linear correlation, while connection-wise normalization can better maintain the linear relationship.

    \subsection{SNN Fine-tuning}
    \subsubsection{Fine-tuning Principle}
      After converting our Spiking-UNet using the aforementioned procedure, we achieve satisfactory results in image segmentation and image denoising tasks. However, the long time window required by the converted network poses computational challenges, limiting its practicality in real-world applications. To address this issue, we employ a training method to fine-tune the converted network and achieve desirable performance in a short time window.

      The BPTT algorithm is a commonly used training algorithm for SNNs. In addition, recent methods such as STBP \cite{wu2018spatio} and ASF-BP \cite{wu2021training} have been proposed. The STBP utilizes the BPTT method through the definition of an approximated derivative, which requires a substantial amount of memory to record the gradient. In our deep Spiking-UNet, the STBP method would cause the GPU memory to overflow. ASF-BP, originally designed for image classification, focuses on training SNNs using the accumulated input and output of an LIF spiking neuron. Compared to STBP, ASF-BP is more computationally efficient as it calculates the gradient only once and requires fewer computational resources. 
      
      Motivated by the concept of accumulated spiking flow in ASF-BP \cite{wu2021training}, we propose a training method for the converted Spiking-UNet model in image segmentation and image denoising tasks. The spiking flow of the multi-threshold spiking neuron is defined as the sum of spikes weighted by the corresponding thresholds. Mathematically, we define the input and output spiking flow as follows:
      \begin{align}
          I_{in}^l &= (\hat{W}^{l})^TI_{out}^{l-1} + \hat{b}^{l}\\
          I_{out}^l &= \begin{cases}
          \sum_{t,i} V_{th,i}s_{i}(t), \quad \text{if} \;\;I_{in}^l > 0 \\
          0\quad \quad \quad \quad \;\;\;\;, \quad \text{otherwise}
          \end{cases}
      \end{align}
    \begin{algorithm}[!t]
      \label{alg:1}
      \caption{Training Spiking-UNet}
      \begin{algorithmic}[1]
          \REQUIRE Parameters and thresholds of a converted Spiking-UNet model $\hat{W}^l, \hat{b}^l, V_{th,i}^l, l=1,2,...,L$; a batch from training dataset $(\textbf{x}, \textbf{y})$
          \ENSURE Update weights and bias
          \\
          \textbf{Forward}
          \STATE $I_{in}^l, I_{out}^l, p_i^l, C_{t}^{l}, s_{t,i}^{l}(t=1,2,...,T)\leftarrow 0$
          \\
          \textbf{Static Coding}
          \STATE $C_{1\cdots T} ^1 \leftarrow \sum_i (\hat{W}^1)^Tx_i + \hat{b}^1$
          \FOR{$t=1$ to $T$}
              \FOR{$l=1$ to $L$}
                  \IF{$l\;not\;equal\;1$}
                      \STATE $C_{t}^{l} \leftarrow \sum_i (\hat{W}^l)^T(V_{th,i}s_{t,i}^{l-1}) + \hat{b}^l$ 
                  \ENDIF
                  \STATE $(s_{t,i}^l, V_{t}^l) \leftarrow Update(s_{t-1,i}^l,C_{t}^l,V_{t-1}^l, V_{th,i}^l)$
                  \STATE $I_{in}^{l} \leftarrow I_{in}^l + C_{t}^{l}$, $I_{out}^l \leftarrow I_{out}^l + \sum_i V_{th,i}s_{t,i}^{l}$
              \ENDFOR
          \ENDFOR
          \FOR{$l=1$ to $L$}
              \STATE $p^l[I_{in}^l > 0] \leftarrow 1$
          \ENDFOR
          \\ \textbf{Loss Calculation}
          \STATE $L\leftarrow \mathcal{L}(y, I_{in}^L)$
          \\ \textbf{Backward}
          \STATE Calculate
          $\frac{\partial L}{\partial \hat{W}^L}$, $\frac{\partial L}{\partial \hat{b}^L}$,$\frac{\partial L}{\partial I_{out}^{L-1}}$, $\frac{\partial L}{\partial \hat{W}^{L-1}}$, $\frac{\partial L}{\partial b^{L-1}}$
          \FOR{$l=L-2$ to $1$}
          \STATE $\frac{\partial L}{\partial I_{out}^l} \leftarrow \frac{\partial L}{\partial I_{out}^{l+1}}p^{l+1}\hat{W}^{l+1}$
          \STATE $\frac{\partial L }{\partial \hat{W}^{l}}\leftarrow \frac{\partial L}{\partial I_{out}^l}p^{l}I_{out}^{l-1}$,\;$\frac{\partial L }{\partial \hat{b}^{l}}\leftarrow \frac{\partial L}{\partial I_{out}^l}p^{l}$
          \ENDFOR
          \\
          \STATE \textbf{Update} $\hat{W}$ \textbf{and} $\hat{b}$
      \end{algorithmic}
  \end{algorithm}  
    \subsubsection{Derivation of Gradients}
      For image segmentation, we use the spiking flow $I_{in}^L$ as the prediction in the output layer, which is then used to calculate the categorical probabilities. We fine-tune the converted SNN model using the categorical cross-entropy loss, which is also utilized in U-Net. The loss is defined as:
      \begin{equation}
          \mathcal{L}_{seg} = -\frac{1}{N^L}\sum_{i=1}^{N^L}\sum_{j=1}^My_{ij} log(\hat{y}_{ij})
      \end{equation}
      where $M$, $N^L$, $y_{ij}$ and $\hat{y}_{ij}$ are the number of categories, the number of neurons in the layer $L$, ground-truth categorical label, and predicted categorical probability of the $i^{th}$ output neuron for the $j^{th}$ category, respectively. 
      
      For the image denoising task, the output of our Spiking-UNet is the noise map, which is utilized for noise removal. The output is obtained from the accumulated spiking flow $I_{in}^L$, which is larger than the noise range. To handle this, we average the spiking flow $I_{in}^L$ over time as an approximation. We optimize the Spiking-UNet using the mean absolute error (MAE) loss between the average spiking flow and the ground truth noise map. The loss is defined as:
      \begin{equation}
          \mathcal{L}_{den} = \frac{1}{N^L}\sum_{x}\sum_{y}|p(x,y)- \frac{I_{in}^L(x,y)}{T}|
      \end{equation}
      where $N^L$, $p(x, y)$ and $\frac{I_{in}^L(x,y)}{T}$ are the number of neurons in the layer $L$, the pixel values of the ground truth and average spiking flow of position $(x, y)$, respectively. Based on the loss and two introduced variables $I_{in}$ and $I_{out}$, we can calculate the gradients of the weight and bias. The whole training procedure is depicted in Algorithm 1. 
\section{Experiments and Results}
  \subsection{Datasets and Implementation}  
    \subsubsection{Segmentation datasets}      
      To evaluate the performance of Spiking-UNet in image segmentation, we choose three datasets: DRIVE \cite{staal2004ridge}, EM segmentation \cite{cardona2010integrated}, and CamSeq01 \cite{fauqueur2007assisted}.
      \begin{itemize}
          \item The DRIVE dataset contains 20 labeled retinal images of size $565 \times 584$ for blood vessel segmentation.
          \item The EM segmentation dataset consists of 30 electron microscopy (EM) images of size $512 \times 512$, which are used for segmenting neuron structures.
          \item The CamSeq01 dataset comprises 101 high-resolution color images of size $960 \times 720$ captured by cameras in an automated driving scenario, with 32 object classes for segmentation.
      \end{itemize}
    \subsubsection{Denoising datasets}
      To assess the effectiveness of Spiking-UNet in image denoising, we train the model on the BSD200 dataset \cite{martin2001database} and evaluate it on the BSD68 and CBSD68 datasets \cite{martin2001database}. 
      \begin{itemize}
          \item The BSD68 dataset consists of 68 grayscale images of size $481 \times 321$ from the Berkeley segmentation dataset. The CBSD68 dataset is the color version of BSD68.
          \item The BSD200 dataset comprises 200 color images of size $481 \times 321$ from the Berkeley segmentation dataset, covering a variety of image types and objects.
      \end{itemize}

  \subsection{Implementation Details}
    \subsubsection{Image Segmentation}
      \label{subsub:image_seg}
      For image processing on the DRIVE dataset, we first use Contrast Limited Adaptive Histogram Equalization (CLAHE) and gamma adjustment operations. Subsequently, we convert these processed images into grayscale. We then partition images into non-overlapping patches of size $48 \times 48$. For the EM segmentation dataset, we deal with input images of size $512 \times 512$, augmented with horizontal and vertical flipping. For the CamSeq01 dataset, we resize the images to a standard dimension of $256 \times 256$ using the bicubic interpolation. At the same, we utilize the nearest interpolation to resize the labels.
      
      Our implementation of the Spiking Fully Convolutional Network (Spiking-FCN) is based on the approach outlined in \cite{kim2022beyond}. We train this model for 100 epochs with a learning rate of $1e-3$. We retain all other settings from the original research \cite{kim2022beyond}.

      For each dataset, we retrain the U-Net model with 200 epochs. We initialize the learning rate at $1.0 \times 10^{-3}$. Batch sizes vary according to the datasets: 8 for DRIVE and CamSeq01, and 1 for EM. We employ the RMSprop optimizer to adjust the learning rate.
        
      We directly train a Spiking-UNet model using multi-level (ML) \cite{feng2022multi}, real spike (RS) \cite{guo2022real}, and multi-threshold (MT) spiking models. We set the epoch number to 100 and the learning rate to $1e-3$, using the Adam optimizer to adjust the learning rate. For the fine-tuning process, we initially use layer-wise and connection-wise weight normalization to adjust weight and bias scales, with the normalization factors of the first layer's input set to one. In our experiments, we set short time windows to 10, 50, and 20 steps on the three datasets, respectively. After converting from the pre-trained U-Net model, we employ our training method to fine-tune the model. For this fine-tuning phase, we use a learning rate of $1.0 \times 10^{-6}$ across all three datasets, and set the epoch number to 100 on the DRIVE, EM, and CamSeq01 datasets.

    \subsubsection{Image Denoising}
      For training purposes, we make use of the BSD200 dataset, where the noise level $\sigma$ ranges from 0 to 55. To augment the data, we employ random cropping techniques to produce patches of $192 \times 256$ pixels. These patches are separately converted into grayscale and color images. The training process operates with a batch size of 8, and we employ the Adam optimizer with an initial learning rate of $1.0 \times 10^{-5}$. Furthermore, the training period spans 400 epochs and incorporates an early stopping strategy to prevent overfitting.

      We directly train a Spiking-UNet model using ML, RS, and MT spiking models over 100 epochs. For the fine-tuning phase, we convert the pre-trained denoising models into their spiking versions and evaluate them using the Set12, BSD68, and CBSD68 datasets. During this evaluation, we add noise at three different levels ($\sigma \in \{ 15,25,50 \}$) to the test images. These noise levels are indicated as $\sigma_{15}$, $\sigma_{25}$, and $\sigma_{50}$, respectively. The output of the Spiking-UNet, treated as the noise map, is the average of the accumulated membrane potentials. The duration of short time windows varies between the datasets, set at 10 and 20 for the BSD68 and CBSD68 datasets, respectively. During the fine-tuning stage, the Spiking-UNet operates with 10/20 time steps for the gray/color versions derived from the BSD200 dataset. Across all datasets, we set the epoch number as 100 during the fine-tuning phase.

  \subsection{Experimental Results}
    \subsubsection{Threshold selection}
    \begin{table}[t]
        \centering
       
        \begin{tabular}{ccccc}
            \toprule
            \multirow{2}{*}{Number} & \multirow{2}{*}{Threshold} & CamSeq01 & \multicolumn{2}{c}{BSD ($\sigma_{25}$)}  \\
             \cmidrule(lr){3-3} \cmidrule(lr){4-5} 
            & & mIoU & PSNR & SSIM \\
            \hline
            1      & 2.0                                & 0.214  & 21.41 & 0.414 \\
            1      & 1.0                                & 0.199  & 26.02 & 0.625 \\
            1      & 0.5                                & 0.306  & 25.84 & 0.613 \\
            2      & 1.0, 0.5                           & 0.325  & 28.17 & 0.760 \\
            2      & 2.0, 1.0                           & 0.363  & 27.16 & 0.706 \\
            3      & 2.0, 1.0, 0.5                      & 0.412  & 28.64 & 0.795 \\
            4      & 2.0, 1.0, 0.5, 0.25                & 0.651  & 28.72 & 0.802 \\
            5      & 2.0, 1.0, 0.5, 0.25, 0.125         & 0.651  & 28.71 & 0.802 \\
            6 & 2.0, 1.0, 0.5, 0.25, 0.125, 0.0625 & 0.650 & 28.71 & 0.804 \\
            \bottomrule 
        \end{tabular}
        \caption{Threshold number and threshold selection for MT spiking neuron using connection-wise normalization on the CamSeq01 and BSD datasets.}
        \label{tab:threshold}
      \end{table}
     For the MT spiking neuron, the number and values of thresholds have a significant impact on the accuracy of the Spiking-UNet model. Table \ref{tab:threshold} examines different combinations of thresholds and their corresponding performance on CamSeq01 and BSD datasets. Initially, we investigate the performance of the IF spiking neuron model with a single threshold, with various threshold values between 2.0, 1.0, and 0.5. We find that the best performance was achieved with a threshold value of 0.5 and 1.0 for CamSeq01 and BSD datasets, respectively. To explore the impact of increasing the number of thresholds, we extend our analysis to MT spiking neurons with three, four, five and six thresholds. The performance of the model continues to improve as the number of thresholds increases, reaching the highest accuracy with a four-threshold configuration (2.0, 1.0, 0.5, 0.25). Interestingly, adding the fifth and sixth thresholds does not result in any noticeable improvement in accuracy compared to the four-threshold configuration. Based on these findings, we select the 4-threshold combination for the MT spiking neurons in our experiments as it achieves the optimal performance.
     
    \subsubsection{Image Segmentation}
     \begin{table*}[t]
      \centering
     
      \resizebox{\linewidth}{!}{
          \begin{tabular}{c|c|c|c|c|c|c|c|c|c|c}
              \toprule
              \multirow{4}{*}{Dataset} & \multirow{4}{*}{T} & \multirow{4}{*}{Metric} & \multirow{4}{*}{U-Net} & \multirow{4}{*}{\shortstack{Spiking-FCN \cite{kim2022beyond}}}  & \multicolumn{2}{c|}{\multirow{2}{*}{ML \cite{feng2022multi}}} & \multicolumn{2}{c|}{\multirow{2}{*}{RS \cite{guo2022real}}} & \multicolumn{2}{c}{\multirow{2}{*}{MT (Ours)}} \\
              & & & & & \multicolumn{2}{c|}{} & \multicolumn{2}{c|}{} & \multicolumn{2}{c}{}\\
              \cline{6-11}
              & & & & & \multirow{2}{*}{DT} & \multirow{2}{*}{FT} & \multirow{2}{*}{DT} & \multirow{2}{*}{FT} & \multirow{2}{*}{DT} & \multirow{2}{*}{FT} \\
              & & & & & & & & & & \\
              \hline
              \multirow{3}{*}{DRIVE} & \multirow{3}{*}{10} & F1 & 0.806 & 0.710 & 0.048 & 0.740 & 0.303 & 0.771 & 0.754 & \textbf{0.794} \\
              & & JS & 0.675 & 0.550 & 0.024 & 0.587 & 0.179 & 0.628 & 0.605 & \textbf{0.658} \\
              & & ACC & 0.970 & 0.960 & 0.921 & 0.960 & 0.668 & 0.967 & 0.965 & \textbf{0.970} \\
              \hline
              \multirow{3}{*}{EM} & \multirow{3}{*}{50} & F1 & 0.948 & - & 0.895 & 0.914 & 0.887 & 0.942 & 0.897 & \textbf{0.947} \\
              & & JS & 0.902 & - & 0.811 & 0.842 & 0.798 & 0.890 & 0.813 & \textbf{0.900} \\
              & & ACC & 0.918 & - & 0.823 & 0.865 & 0.810 & 0.909 & 0.828 & \textbf{0.917} \\
              \hline
              \multirow{2}{*}{CamSeq01} & \multirow{2}{*}{20} & \multirow{2}{*}{mIoU} & \multirow{2}{*}{0.620} & \multirow{2}{*}{0.313} & \multirow{2}{*}{0.032} & \multirow{2}{*}{0.165} & \multirow{2}{*}{-} & \multirow{2}{*}{0.208} & \multirow{2}{*}{0.081} & \multirow{2}{*}{\textbf{0.651}}\\
              & & & & & & & & & & \\
              \bottomrule 
          \end{tabular}
          
    }
     \caption{The quantitative segmentation results of the U-Net models, Spiking-FCN, different neuron models using direct training and fine-tuning on DRIVE, EM and CamSeq01 datasets. ML, RS and MT represent Multi-Level spiking neuron, Real Spike spiking neuron and Multi-Threshold spiking neuron, respectively. `DT' and `FT' represent direct training and fine-tuning, respectively. `T' represents the number of time steps.}
    \label{tab:seg}
  \end{table*}
  \begin{figure*}[t]
    \centering
    \includegraphics[width=1.0\textwidth]{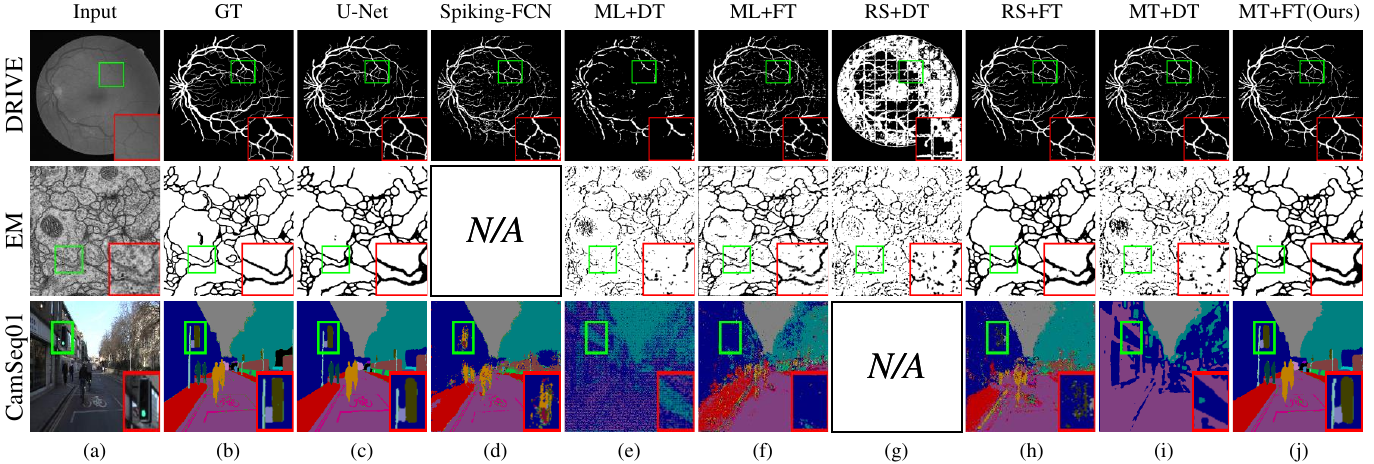}
    \caption{Qualitative segmentation results on images from the DRIVE, EM, and CamSeq01 datasets. (a) Input. (b) Ground Truth (GT). Segmentation results with (c) U-Net. (d) Spiking-FCN. (e, f) Multi-Level (ML) spiking neuron with direct training and fine-tuning, respectively. (g, h) Real Spike (RS) spiking neuron with direct training and fine-tuning, respectively. (i, j) Multi-Threshold (MT) spiking neuron with direct training and fine-tuning, respectively. `DT' and `FT' represent direct training and fine-tuning, respectively.}
    \label{fig:seg}
  \end{figure*}
      In segmentation tasks, there are four metrics commonly used to measure the performance, which are Jaccard Similarity (JS), F1 Score (F1), Pixel-wise Accuracy (ACC) and Mean Intersection-Over-Union (mIoU). JS, also known as Intersection-Over-Union, is a statistic to measure the similarity between the segmentation result and ground truth. F1 measures the segmentation by calculating the harmonic mean of the precision and recall. ACC depicts the proportion of correct segmented pixels. mIoU calculates the Intersection-Over-Union for each semantic class and then averages over classes. For binary segmentation, the former three metrics are utilized. For multi-class segmentation, mIoU is adopted.
              
      Table \ref{tab:seg} shows the quantitative segmentation results of the U-Net model, the Spiking-FCN model and the Spiking-UNet models with different spiking neurons and training strategies on three datasets. In terms of direct training, the Spiking-UNet with the MT spiking neuron consistently outperforms those with the ML and RS spiking neuron models on all datasets. The MT spiking neuron exhibits better information representation and captures spiking activity more accurately, leading to improved segmentation results. After fine-tuning, our Spiking-UNet with the MT spiking neuron continues to outperform other methods at the same settings, and even achieves comparable results to the U-Net model. Our fine-tuning strategies effectively refine the converted Spiking-UNet models, leveraging the strengths of spiking neurons while reducing the performance gap between spiking and traditional neural networks. This demonstrates the effectiveness of the proposed MT spiking neuron and the fine-tuning strategy in achieving high-quality segmentation results.
      
      Figure \ref{fig:seg} illustrates the qualitative segmentation results of the U-Net model, the Spiking-FCN model, and Spiking-UNet with different spiking neurons using direct training and fine-tuning on three datasets in a short time window. When examining the results of direct training, it is evident that Spiking-FCN and the Spiking-UNet with RS spiking neurons fail to generate satisfactory segmentation results on the EM and CamSeq01 datasets, respectively. In fact, across all datasets, the segmentation results obtained from direct training methods are generally suboptimal, indicating the challenges in effectively training spiking neural networks for image segmentation tasks. However, upon closer examination of the results from fine-tuning strategies, a significant improvement in the segmentation performance can be observed. Fine-tuning strategies refine and adjust the converted spiking neural networks, leading to enhanced segmentation accuracy. Notably, our proposed method, which utilizes the MT spiking neuron and undergoes fine-tuning, consistently achieves results that are comparable to those of the U-Net models. This demonstrates the effectiveness of our approach in bridging the gap between the traditional U-Net model and spiking neural networks.
      \begin{table*}[t]
        \centering
      
        \resizebox{\linewidth}{!}{
            \begin{tabular}{c|c|c|c|c|c|c|c|c|c|c}
                \toprule
                \multirow{4}{*}{Dataset} & \multirow{4}{*}{T} & \multirow{4}{*}{Metric} & \multirow{4}{*}{$\sigma$} & \multirow{4}{*}{U-Net} & \multicolumn{2}{c|}{\multirow{2}{*}{ML \cite{feng2022multi}}} & \multicolumn{2}{c|}{\multirow{2}{*}{RS \cite{guo2022real}}} & \multicolumn{2}{c}{\multirow{2}{*}{MT (Ours)}} \\
                & & & & & \multicolumn{2}{c|}{} & \multicolumn{2}{c|}{} & \multicolumn{2}{c}{}\\
                \cline{6-11}
                & & & & & \multirow{2}{*}{DT} & \multirow{2}{*}{FT} & \multirow{2}{*}{DT} & \multirow{2}{*}{FT} & \multirow{2}{*}{DT} & \multirow{2}{*}{FT} \\
                & & & & & & & & & & \\
                \hline
                \multirow{6}{*}{BSD68} & \multirow{6}{*}{10} & \multirow{3}{*}{PSNR} & 15 & 31.02 & 23.92 & 28.47 & 24.57 & 28.51 & 25.75 & \textbf{31.08} \\
                & & & 25 & 28.61 & 22.71 & 25.98 & 23.73 & 26.13 & 25.06 & \textbf{28.72}\\
                & & & 50 & 25.72 & 18.92 & 21.80 & 19.22 & 21.96 & 22.76 & \textbf{25.86}\\
                \cline{3-11}
                & & \multirow{3}{*}{SSIM} & 15 & 0.868 & 0.544 & 0.740 & 0.572 & 0.737 & 0.757 & \textbf{0.869} \\
                & & & 25 & 0.799 & 0.480 & 0.615 & 0.534 & 0.616 & 0.694 & \textbf{0.802}\\
                & & & 50 & 0.684 & 0.308 & 0.375 & 0.324 & 0.396 & 0.524 & \textbf{0.692}\\
                \hline
                \multirow{6}{*}{CBSD68} & \multirow{6}{*}{20} & \multirow{3}{*}{PSNR} & 15 & 33.32 & 23.77 & 26.31 & 26.10 & 27.57 & 27.84 & \textbf{33.14}\\
                & & & 25 & 30.59 & 21.27 & 23.87 & 23.55 & 24.74 & 26.44 & \textbf{30.57}\\
                & & & 50 & 27.14 & 16.41 & 19.74 & 18.78 & 19.23 & 22.86 & \textbf{27.24}\\
                \cline{3-11}
                & & \multirow{3}{*}{SSIM} & 15 & 0.920 & 0.538 & 0.634 & 0.665 & 0.685 & 0.828 & \textbf{0.913}\\
                & & & 25 & 0.864 & 0.424 & 0.515 & 0.537 & 0.547 & 0.739 & \textbf{0.858}\\
                & & & 50 & 0.749 & 0.236 & 0.333 & 0.325 & 0.328 & 0.539 & \textbf{0.753}\\
               
                \bottomrule
            \end{tabular}
        }
          \caption{The quantitative denoising results of the U-Net models, different neuron models using direct training and fine-tuning on BSD68 and CBSD68 datasets. ML, RS and MT represent Multi-Level spiking neuron, Real Spike spiking neuron and Multi-Threshold spiking neuron, respectively. `DT' and `FT' represent direct training and fine-tuning, respectively. `T' represents the number of time steps.}
        \label{tab:den}
      \end{table*}

    \subsubsection{Image Denoising}
    
      Different from image segmentation, image denoising predicts continuous values. Due to discrete spikes, the spiking neuron has difficulty predicting the accurate values, especially in a short time window. We evaluate our Spiking-UNet on the Set12, BSD68, and CBSD68 datasets using two metrics: PSNR and SSIM. The Peak signal-to-noise ratio (PSNR) is commonly used to measure the quality of reconstructed images through the ratio of the maximum possible pixel value of the image to the mean squared error computed from two images. The Structural Similarity Index (SSIM) is a perceptual metric that quantifies image quality degradation based on visible structures in the image.
      
      Table \ref{tab:den} provides the quantitative evaluation results of pre-trained U-Net models and their corresponding Spiking-UNet models, equipped with different spiking neuron models, on the BSD68 and CBSD68 datasets, as measured by PSNR and SSIM metrics. When it comes to direct training, our Spiking-UNet with MT spiking neurons outperforms those with ML and RS spiking neurons. After fine-tuning, the Spiking-UNet model with MT spiking neurons achieves results comparable to the U-Net model, demonstrating the effectiveness of this neuron type and fine-tuning strategy in image denoising tasks.
       \begin{figure*}[t]
          \centering
          \includegraphics[width=1.0\textwidth]{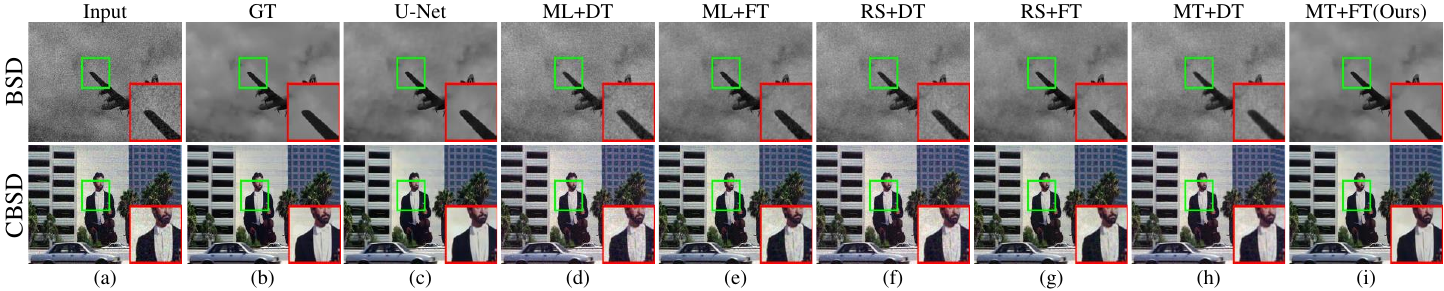}
          \caption{Qualitative denoising results on images from the BSD68, and CBSD68 datasets. (a) Input (b) Ground Truth (GT). Denoising results with (c) U-Net. (d, e) Multi-Level (ML) spiking neuron with direct training and fine-tuning, respectively. (f, g) Real Spike (RS) spiking neuron with direct training and fine-tuning, respectively. (h, i) Multi-Threshold (MT) spiking neuron with direct training and fine-tuning, respectively.}
          \label{fig:den}
      \end{figure*}
      
      Figure \ref{fig:den} presents the qualitative image denoising results of our Spiking-UNet when subjected to noise level $\sigma_{25}$. For a more detailed analysis, we have marked regions of interest with green boxes on each test image and enlarged these regions within red boxes. With direct training, the Spiking-UNet employing MT spiking neurons shows superior noise removal as compared to the ML and RS spiking neurons. However, Spiking-UNet using direct training still shows room for improvement in noise removal efficiency. Upon applying our fine-tuning strategy, we observed enhancements across all spiking neuron model results. Notably, the Spiking-UNet with MT spiking neurons yields smooth results, demonstrating performance on par with traditional U-Net models.
  
  \subsection{Ablation Study}
    \subsubsection{Effect of Normalization, MT spiking neuron and Fine-tuning}
    \begin{table}[t]
      \centering 
      \begin{tabular}{cccc|cc|ccc}
          \toprule
          \multicolumn{4}{c|}{Methods} &
          \multicolumn{2}{c|}{CamSeq01} & \multicolumn{3}{c}{BSD ($\sigma_{25}$)} \\
          \hline
          LW & CW & MT & FT & T & mIoU  & T & PSNR & SSIM \\
          \hline
          $\times$   & \checkmark & $\times$   & $\times$ & 1000 & 0.592 & 
           200 & 28.61 & 0.802\\
          
          $\times$   & \checkmark & \checkmark & $\times$ & 1000 & 0.623 & 
           200 & 28.67 & 0.804\\
          
          $\times$   & \checkmark & $\times$   & $\times$ & 20 & 0.002 & 
           10 & 21.51 & 0.428\\
          
          $\times$   & \checkmark & \checkmark & $\times$ & 20 & 0.272 & 
           10 & 23.37 & 0.511\\
          \hline
          \checkmark & $\times$   & $\times$   & \checkmark & 20 & 0.144 & 
           10 & 25.61 & 0.593\\
          
          \checkmark & $\times$   & \checkmark & \checkmark & 20 & 0.199 & 
           10 & 28.53 & 0.790 \\
          
          $\times$   & \checkmark & $\times$   & \checkmark & 20 & 0.603 &
           10 & 26.40 & 0.639 \\
          
          $\times$   & \checkmark & \checkmark & \checkmark & 20 & \textbf{0.651} & 
           10 & \textbf{28.72} & \textbf{0.802}\\
          \bottomrule
      \end{tabular}
      \caption{Ablation Study of normalization, MT spiking neuron, Fine-tuning strategy on CamSeq01 and BSD datasets. `LW' and `CW' represent the layer-wise and connection-wise normalization methods, respectively. `T' represents the number of time steps.}
      \label{tab:ablation_study}
    \end{table}
      Table \ref{tab:ablation_study} presents an ablation study investigating the impact of various components on the performance of the Spiking-UNet model using the CamSeq01 and BSD datasets. We examine components including normalization, the MT spiking neuron, and fine-tuning strategies. For the converted Spiking-UNet models without fine-tuning, it is observed that the inclusion of the MT spiking neuron improves the performance at both 1000/200 and 20/10 time steps for CamSeq01 and BSD datasets, respectively. Specifically, the mIoU increases from 0.592 to 0.623 when the MT spiking neuron is introduced. This indicates that the MT spiking neuron enhances the accuracy in the absence of fine-tuning. Furthermore, the results demonstrate that when fine-tuning is applied to the converted Spiking-UNet models, the utilization of connection-wise normalization leads to superior performance compared to layer-wise normalization. Specifically, when combined with the MT spiking neuron, the connection-wise normalization achieves the highest performance, resulting in mIoU 0.651 for CamSeq01, PSNR/SSIM 28.72/0.802 for BSD ($\sigma_{25}$). This finding highlights the effectiveness of both the MT spiking neuron and the fine-tuning strategy with the connection-wise normalization in significantly improving the segmentation and denoising results.

    \subsubsection{Effect of Timesteps}
     \begin{table}[t]
        \centering
        \begin{tabular}{c|c|cccccccccc}
            \toprule
            \multicolumn{2}{c|}{Timesteps}                   & 4  & 8  & 10 & 16 & 20 & 24 & 28 \\
            \hline
            CamSeq01 & mIoU & 0.251 & 0.415 & 0.505 & 0.606 & 0.651 & 0.641 & 0.649 \\
            \hline
            \multirow{2}{*}{BSD ($\sigma_{25}$)} & PSNR         & 28.04 & 28.58 & 28.72 & 28.72 & 28.71 & 28.73 & 28.74\\
                                 & SSIM         & 0.745 & 0.790 & 0.802 & 0.803 & 0.803 & 0.804 & 0.804 \\
        \bottomrule            
        \end{tabular}
        \caption{Comparison of various timesteps on the CamSeq01 and BSD datasets.}
        \label{tab:Timesteps}
    \end{table} 
    Table \ref{tab:Timesteps} presents the performance with different timesteps on the CamSeq01 and BSD datasets. For the CamSeq01 dataset, mIoU gets better up to 20 steps and then it stays around 0.651. For the BSD dataset, the PSNR and SSIM get better as the steps go up. After 10 steps, these scores do not change much and stay around 28.72 for PSNR and 0.802 for SSIM.
    \subsubsection{Energy Efficiency}
        \begin{table}[t]
            \centering
            \begin{tabular}{ccccc}
                \hline
                Model                   & FLOPs    & Mem. (J) & Ops. (J) & Total (J) \\
                \hline
                U-Net                  & 6.55E+10 & 5.19E-2 & 3.01E-1 & 3.53E-1 \\
                Spiking-FCN            & 1.90E+7  & 7.58E-4 & 5.57E-4 & 1.32E-3 \\
                Spiking-UNet (ML)      & 7.11E+7  & 1.78E-3 & 6.04E-4 & 2.38E-3 \\
                Spiking-UNet (RS)      & 6.96E+7  & 1.11E-3 & 6.03E-4 & 1.71E-3 \\
                Spiking-UNet (MT)      & 2.92E+8  & 1.24E-3 & 6.98E-4 & 1.94E-3 \\
                \hline                           
            \end{tabular}
            \caption{Comparison of Energy Consumption between U-Net model, Spiking-FCN model and corresponding Spiking-UNet models using different spiking neurons with connection-wise normalization after fine-tuning on the CamSeq01 dataset.}
            \label{tab:energy}
        \end{table} 
       Energy consumption is a critical factor when evaluating the cost of model inference.  To provide a comprehensive assessment of the energy consumption of the Spiking-UNet model, we consider the energy cost of each operation and memory access based on the CamSeq01 dataset. The energy cost for Multiply-Accumulate (MAC) operations is estimated to be 4.6pJ, while Arithmetic and Comparison (AC) operations have an energy cost of 0.9pJ, as suggested by Horowitz \emph{et al.} \cite{horowitz20141}. We utilize the method \cite{lemaire2022analytical} to calculate the energy cost of accessing SRAM memory. Following the method described in \cite{qu2023spiking}, we define the FLOPs used to compute the energy consumption of the Spiking-UNet as

      \begin{equation}
          FLOPs = \#OP_{spikes} + \#OP_{input\;layer}
      \end{equation}
       where $\#OP_{spikes}$ represents the total spike counts of SNN and $\#OP_{input\;layer}$ denotes the FLOPs of ANN in the first layer using direct coding. Table \ref{tab:energy} presents the quantitative results of energy consumption for the U-Net model, Spiking-FCN model and corresponding Spiking-UNet models using different spiking neurons with connection-wise normalization after fine-tuning on the CamSeq01 dataset. Our Spiking-UNet achieves higher mIoU while consuming two orders of magnitude less energy compared to the U-Net model, showcasing its advantage in terms of energy consumption.

\section{Conclusion and Discussion}
    In this paper, we introduce Spiking-UNet, a deep SNN for image processing, specifically designed for image segmentation and image denoising tasks. To achieve an efficient Spiking-UNet, we need to address the challenges of high-fidelity information propagation and the development of an effective training strategy. To overcome these challenges, we propose multi-threshold spiking neurons to enhance high-fidelity information transmission within the network. Furthermore, we utilize a conversion and fine-tuning pipeline that leverages pre-trained U-Net models, which ensures the effective training of our Spiking-UNet. We address inconsistent spiking rates caused by the significant variability in data distribution of skip connections through the application of a connection-wise normalization method during the conversion process. Additionally, we introduce a training method based on the spiking flow, which enables fine-tuning of the converted models while reducing the number of time steps required for inference. Experimental results demonstrate that our Spiking-UNet not only achieves comparable performance to the non-spiking U-Net model but also outperforms existing SNN methods for image segmentation and denoising tasks. Notably, our approach significantly reduces inference time by approximately 90\% compared to the Spiking-UNet model without our fine-tuning.     
    
Our research still has several limitations. As a preliminary exploration, we evaluate our Spiking-UNet on traditional and relatively small datasets for quick evaluation. In the future, Spiking-UNet will be tested on more newer and larger datasets. In addition, we only utilize Spiking-UNet on two image processing tasks. We will extend the application of Spiking-UNet, such as image super-resolution. Furthermore, we will explore the deployment of Spiking-UNet on neuromorphic chips to validate its effectiveness in real world.

\section{Acknowledgements}
This work was supported in part by the National Natural Science Foundation of China under Grant 62021001.

\appendix
\section{}
\begin{mytheorem}
    Consider a multi-threshold neuron model with a membrane voltage $V$ that follows a uniform distribution in the range $[0, 1]$ ($V \sim U[0, 1]$). Let $N$ thresholds be denoted as $V_{th,1}, V_{th,2}, ..., V_{th,N}$, subject to the constraints $V_{th,1} > V_{th,2} > ... > V_{th,N-1} > V_{th,N}$ and $\sum_i V_{th,i} \leq 1$.
    
    Suppose the multi-threshold spiking neuron can fire only one spike for each threshold. Then the membrane voltage $V$ can be expressed as $V = \alpha_1 V_{th,1} + \alpha_2 V_{th,2} + ... + \alpha_N V_{th, N} + r(V)$, where $\alpha_i \in \{0, 1\}$ indicates whether a spike is fired at each threshold, and $r(V)$ represents the errors between the actual membrane voltage $V$ and the output weighted sum of spikes.
    
    The optimal series of thresholds that minimizes the error expectation satisfies $V_{th,i+1}=V_{th,i}/{2}$ and $V_{th, N} = 1/{2^N}$, yielding a minimal error expectation of $r^* = 1/{2^{N+1}}$.
\end{mytheorem}
\begin{myproof}
    Denote $\alpha_1 V_{th,1} + \alpha_2 V_{th,2} + ... + \alpha_N V_{th, N}$ as a binary sequence $(\alpha_1, \alpha_2, ..., \alpha_N)_2$. N thresholds divide the range $[0, 1]$ into $2^N$ segments. Let $a_i$ represent the left endpoint of the $i^{th} (i \in \{1, \cdots, 2^N \})$ segment of $(\alpha_1, \alpha_2, ..., \alpha_N)_2$ to $((\alpha_1, \alpha_2, ..., \alpha_N)_2+1)_2$. In particular, we have $a_1 = 0$ and define $a_{2^N+1} = 1$. When $V$ is in the range of $a_i$ and  $a_{i+1}$, the error $r(V)$ is equal to $V - a_i$. Therefore, $r$ is formulated as 
    \begin{equation}
    \begin{aligned}
        r &= \sum_{i=1}^{2^{N}}\int_{a_i}^{a_{i+1}}(V-a_i)dV\\
        &=\sum_{i=1}^{2^{N}}\int_0^{a_{i+1}-a_i} z dz
        = \frac{1}{2}\sum_{i=1}^{2^{N}}{(a_{i+1}-a_i)}^2
    \end{aligned}
    \end{equation}
    Let $b_i=a_{i+1}-a_{i}$. We minimize the error function as follows:
    \begin{equation}
    \mathop{argmin}\limits_{b_i}\frac{1}{2}\sum_{i=1}^{2^{N}}b_i^2, s.t. \sum_{i=1}^{2^{N}} b_i = 1
    \end{equation}
    Using the Cauchy–Schwarz inequality, we obtain that when $b_i = 1/{2^N}$ (for $i={1,2,...,2^N}$), the objective function can achieve the lowest value $1/{2^{N+1}}$. For all $b_i = 1/{2^N}$, the $V_{th,i+1}$ is equal to $V_{th, i}/{2}$ and $V_{th,N} = 1/{2^N}$.
    
    The above theorem provides the standard to determine thresholds for $V \sim [0, 1]$. When the distribution of $V$ changes, for example $V \sim [0,4]$, the solution should be slightly altered.
\end{myproof}
\section{}
\begin{table}[!t]
    \centering
    \footnotesize
    \setlength{\tabcolsep}{4pt}
    \begin{tabular}{c|c|c|c|c|c|c}
       \hline
       Dataset & Metric & ML + FT & RS + FT & MT + FT & \begin{tabular}{@{}c@{}}
        ML + FT \\
        v.s. \\
        MT + FT \\
        p-value
        \end{tabular} & \begin{tabular}{@{}c@{}}
        RS + FT \\
        v.s. \\
        MT + FT \\
        p-value
        \end{tabular}\\
       \hline
       \multirow{3}{*}{DRIVE}  & F1 & 0.740 & 0.771 & 0.794 & 2.62e-7 & 1.63e-2 \\
                           & JS & 0.587 & 0.628 & 0.658 & 2.56e-7 & 1.71e-2 \\
                           & ACC & 0.960 & 0.967 & 0.970 & 1.71e-7 & 5.59e-2 \\
       \hline
       \multirow{3}{*}{EM} & F1 & 0.914 & 0.942 & 0.947 & 3.73e-4 & 1.22e-2 \\
                           & JS & 0.842 & 0.890 & 0.900 & 3.39e-4 & 1.19e-2 \\
                           & ACC & 0.865 & 0.909 & 0.917 & 6.86e-5 & 3.90e-3 \\
        \hline
        CamSeq01 & mIoU & 0.165 & 0.208 & 0.651 & 4.30e-31 & 2.36e-29 \\
       \hline
       \multirow{2}{*}{BSD68 ($\sigma_{25}$)} & PSNR & 25.98 & 26.13 & 28.72 & 6.73e-13 & 1.15e-11 \\ 
                              & SSIM & 0.615 & 0.616 & 0.802 & 4.58e-22 & 4.39e-22 \\
        \hline
        \multirow{2}{*}{CBSD68($\sigma_{25}$)} & PSNR & 23.87 & 24.74 & 30.57 & 4.13e-24 & 4.13e-24 \\
                                & SSIM & 0.515 & 0.547 & 0.858 & 7.02e-24 & 9.97e-24 \\
        \hline
    \end{tabular}
    \caption{The quantitative segmentation results and p-value of different spiking neurons on various datasets. `ML', `RS' and `MT' represent Multi-Level spiking neurons, Real Spike spiking neurons and Multi-Threshold spiking neurons, respectively. }
    \label{tab:p_value}
\end{table}

To evaluate the significant improvement of our methods, we utilize hypothesis testing across different datasets. Initially, the null hypothesis is that the MT + FT method is significantly better than the ML/RS + FT method. Then, we compute the metric of each test sample. We use the Shapiro-Wilk test to determine whether the data of metrics follows a normal distribution. For normally distributed data, we use the t-test; for non-normal data, the Mann-Whitney U test is applied. Finally, this analysis produces the p-values, which indicate our method is statistically significant if the p-values are smaller than the significance level.
Table \ref{tab:p_value} shows the quantitative segmentation results and p-values across different datasets. We set the significance level $\alpha$ as 0.05. It is observed that the MT + FT method is significantly better than the ML + FT method. Besides, in most cases, the MT + FT method shows significant improvement over the RS + FT method, except for the accuracy metric in the DRIVE dataset.

\bibliographystyle{elsarticle-num} 
\bibliography{elsarticle}

\begin{thebibliography}{10}
\expandafter\ifx\csname url\endcsname\relax
  \def\url#1{\texttt{#1}}\fi
\expandafter\ifx\csname urlprefix\endcsname\relax\def\urlprefix{URL }\fi
\expandafter\ifx\csname href\endcsname\relax
  \def\href#1#2{#2} \def\path#1{#1}\fi

\bibitem{ronneberger2015u}
O.~Ronneberger, P.~Fischer, T.~Brox, U-net: Convolutional networks for
  biomedical image segmentation, in: International Conference on Medical Image
  Computing and Computer-Assisted Intervention, Springer, 2015, pp. 234--241.

\bibitem{xiao2020global}
B.~Xiao, B.~Xu, X.~Bi, W.~Li, Global-feature encoding u-net (geu-net) for
  multi-focus image fusion, IEEE Transactions on Image Processing 30 (2020)
  163--175.

\bibitem{zhou2020hierarchical}
S.~Zhou, J.~Wang, J.~Zhang, L.~Wang, D.~Huang, S.~Du, N.~Zheng, Hierarchical
  u-shape attention network for salient object detection, IEEE Transactions on
  Image Processing 29 (2020) 8417--8428.

\bibitem{liu2022video}
T.~Liu, Q.~Meng, J.-J. Huang, A.~Vlontzos, D.~Rueckert, B.~Kainz, Video
  summarization through reinforcement learning with a 3d spatio-temporal u-net,
  IEEE Transactions on Image Processing 31 (2022) 1573--1586.

\bibitem{nazir2021ecsu}
A.~Nazir, M.~N. Cheema, B.~Sheng, P.~Li, H.~Li, G.~Xue, J.~Qin, J.~Kim, D.~D.
  Feng, Ecsu-net: An embedded clustering sliced u-net coupled with fusing
  strategy for efficient intervertebral disc segmentation and classification,
  IEEE Transactions on Image Processing 31 (2021) 880--893.

\bibitem{cciccek20163d}
{\"O}.~{\c{C}}i{\c{c}}ek, A.~Abdulkadir, S.~S. Lienkamp, T.~Brox,
  O.~Ronneberger, 3d u-net: Learning dense volumetric segmentation from sparse
  annotation, in: International Conference on Medical Image Computing and
  Computer-Assisted Intervention, Springer, 2016, pp. 424--432.

\bibitem{sengupta2019going}
A.~Sengupta, Y.~Ye, R.~Wang, C.~Liu, K.~Roy, Going deeper in spiking neural
  networks: Vgg and residual architectures, Frontiers in Neuroscience 13 (2019)
  95.

\bibitem{liao2023convolutional}
X.~Liao, Y.~Wu, Z.~Wang, D.~Wang, H.~Zhang, A convolutional spiking neural
  network with adaptive coding for motor imagery classification, Neurocomputing
  (2023) 126470.

\bibitem{zhang2019tdsnn}
L.~Zhang, S.~Zhou, T.~Zhi, Z.~Du, Y.~Chen, Tdsnn: From deep neural networks to
  deep spike neural networks with temporal-coding, in: Proceedings of the AAAI
  Conference on Artificial Intelligence, Vol.~33, 2019, pp. 1319--1326.

\bibitem{yang2022training}
Q.~Yang, J.~Wu, M.~Zhang, Y.~Chua, X.~Wang, H.~Li, Training spiking neural
  networks with local tandem learning, 36rd Conference on Neural Information
  Processing Systems (2022).

\bibitem{akopyan2015truenorth}
F.~Akopyan, J.~Sawada, A.~Cassidy, R.~Alvarez-Icaza, J.~Arthur, P.~Merolla,
  N.~Imam, Y.~Nakamura, P.~Datta, G.-J. Nam, et~al., Truenorth: Design and tool
  flow of a 65 mw 1 million neuron programmable neurosynaptic chip, IEEE
  Transactions on Computer-Aided Design of Integrated Circuits and Systems
  34~(10) (2015) 1537--1557.

\bibitem{davies2018loihi}
M.~Davies, N.~Srinivasa, T.-H. Lin, G.~Chinya, Y.~Cao, S.~H. Choday, G.~Dimou,
  P.~Joshi, N.~Imam, S.~Jain, et~al., Loihi: A neuromorphic manycore processor
  with on-chip learning, IEEE Micro 38~(1) (2018) 82--99.

\bibitem{pei2019towards}
J.~Pei, L.~Deng, S.~Song, M.~Zhao, Y.~Zhang, S.~Wu, G.~Wang, Z.~Zou, Z.~Wu,
  W.~He, et~al., Towards artificial general intelligence with hybrid tianjic
  chip architecture, Nature 572~(7767) (2019) 106--111.

\bibitem{furber2014spinnaker}
S.~B. Furber, F.~Galluppi, S.~Temple, L.~A. Plana, The spinnaker project,
  Proceedings of the IEEE 102~(5) (2014) 652--665.

\bibitem{fang2021deep}
W.~Fang, Z.~Yu, Y.~Chen, T.~Huang, T.~Masquelier, Y.~Tian, Deep residual
  learning in spiking neural networks, Advances in Neural Information
  Processing Systems 34 (2021) 21056--21069.

\bibitem{feng2022multi}
L.~Feng, Q.~Liu, H.~Tang, D.~Ma, G.~Pan, Multi-level firing with spiking
  ds-resnet: Enabling better and deeper directly-trained spiking neural
  networks, Proceedings of the Thirty-First International Joint Conference on
  Artificial Intelligence (2022).

\bibitem{guo2022real}
Y.~Guo, L.~Zhang, Y.~Chen, X.~Tong, X.~Liu, Y.~Wang, X.~Huang, Z.~Ma, Real
  spike: Learning real-valued spikes for spiking neural networks, in: Computer
  Vision--ECCV 2022: 17th European Conference, Tel Aviv, Israel, October
  23--27, 2022, Proceedings, Part XII, Springer, 2022, pp. 52--68.

\bibitem{fang2021incorporating}
W.~Fang, Z.~Yu, Y.~Chen, T.~Masquelier, T.~Huang, Y.~Tian, Incorporating
  learnable membrane time constant to enhance learning of spiking neural
  networks, in: Proceedings of the IEEE/CVF International Conference on
  Computer Vision, 2021, pp. 2661--2671.

\bibitem{wu2023dynamic}
X.~Wu, Y.~Zhao, Y.~Song, Y.~Jiang, Y.~Bai, X.~Li, Y.~Zhou, X.~Yang, Q.~Hao,
  Dynamic threshold integrate and fire neuron model for low latency spiking
  neural networks, Neurocomputing 544 (2023) 126247.

\bibitem{xu2023ultra}
C.~Xu, Y.~Liu, Y.~Yang, Ultra-low latency spiking neural networks with
  spatio-temporal compression and synaptic convolutional block, Neurocomputing
  (2023) 126485.

\bibitem{chen2022adaptive}
Y.~Chen, Y.~Mai, R.~Feng, J.~Xiao, An adaptive threshold mechanism for accurate
  and efficient deep spiking convolutional neural networks, Neurocomputing 469
  (2022) 189--197.

\bibitem{staal2004ridge}
J.~Staal, M.~D. Abr{\`a}moff, M.~Niemeijer, M.~A. Viergever, B.~Van~Ginneken,
  Ridge-based vessel segmentation in color images of the retina, IEEE
  Transactions on Medical Imaging 23~(4) (2004) 501--509.

\bibitem{cardona2010integrated}
A.~Cardona, S.~Saalfeld, S.~Preibisch, B.~Schmid, A.~Cheng, J.~Pulokas,
  P.~Tomancak, V.~Hartenstein, An integrated micro- and macroarchitectural
  analysis of the drosophila brain by computer-assisted serial section electron
  microscopy, PLoS Biology 8~(10) (2010) e1000502.

\bibitem{fauqueur2007assisted}
J.~Fauqueur, G.~Brostow, R.~Cipolla, Assisted video object labeling by joint
  tracking of regions and keypoints, in: 2007 IEEE 11th International
  Conference on Computer Vision, IEEE, 2007, pp. 1--7.

\bibitem{martin2001database}
D.~Martin, C.~Fowlkes, D.~Tal, J.~Malik, A database of human segmented natural
  images and its application to evaluating segmentation algorithms and
  measuring ecological statistics, in: Proceedings Eighth IEEE International
  Conference on Computer Vision, Vol.~2, IEEE, 2001, pp. 416--423.

\bibitem{abbott1999lapicque}
L.~F. Abbott, Lapicque’s introduction of the integrate-and-fire model neuron
  (1907), Brain Research Bulletin 50~(5-6) (1999) 303--304.

\bibitem{doutsi2021dynamic}
E.~Doutsi, L.~Fillatre, M.~Antonini, P.~Tsakalides, Dynamic image quantization
  using leaky integrate-and-fire neurons, IEEE Transactions on Image Processing
  30 (2021) 4305--4315.

\bibitem{li2022brain}
W.~Li, H.~Chen, J.~Guo, Z.~Zhang, Y.~Wang, Brain-inspired multilayer perceptron
  with spiking neurons, in: Proceedings of the IEEE/CVF Conference on Computer
  Vision and Pattern Recognition, 2022, pp. 783--793.

\bibitem{wu2021liaf}
Z.~Wu, H.~Zhang, Y.~Lin, G.~Li, M.~Wang, Y.~Tang, Liaf-net: Leaky integrate and
  analog fire network for lightweight and efficient spatiotemporal information
  processing, IEEE Transactions on Neural Networks and Learning Systems (2021).

\bibitem{diehl2015fast}
P.~U. Diehl, D.~Neil, J.~Binas, M.~Cook, S.-C. Liu, M.~Pfeiffer,
  Fast-classifying, high-accuracy spiking deep networks through weight and
  threshold balancing, in: 2015 International Joint Conference on Neural
  Networks, IEEE, 2015, pp. 1--8.

\bibitem{rueckauer2017conversion}
B.~Rueckauer, I.-A. Lungu, Y.~Hu, M.~Pfeiffer, S.-C. Liu, Conversion of
  continuous-valued deep networks to efficient event-driven networks for image
  classification, Frontiers in Neuroscience 11 (2017) 682.

\bibitem{yan2021near}
Z.~Yan, J.~Zhou, W.-F. Wong, Near lossless transfer learning for spiking neural
  networks, in: Proceedings of the AAAI Conference on Artificial Intelligence,
  Vol.~35, 2021, pp. 10577--10584.

\bibitem{kim2018spiking}
H.~Kim, S.~Hwang, J.~Park, S.~Yun, J.-H. Lee, B.-G. Park, Spiking neural
  network using synaptic transistors and neuron circuits for pattern
  recognition with noisy images, IEEE Electron Device Letters 39~(4) (2018)
  630--633.

\bibitem{diehl2015unsupervised}
P.~U. Diehl, M.~Cook, Unsupervised learning of digit recognition using
  spike-timing-dependent plasticity, Frontiers in Computational Neuroscience 9
  (2015) 99.

\bibitem{liu2020unsupervised}
Q.~Liu, G.~Pan, H.~Ruan, D.~Xing, Q.~Xu, H.~Tang, Unsupervised aer object
  recognition based on multiscale spatio-temporal features and spiking neurons,
  IEEE Transactions on Neural Networks and Learning Systems 31~(12) (2020)
  5300--5311.

\bibitem{liu2019deep}
D.~Liu, N.~Bellotto, S.~Yue, Deep spiking neural network for video-based
  disguise face recognition based on dynamic facial movements, IEEE
  Transactions on Neural Networks and Learning Systems 31~(6) (2019)
  1843--1855.

\bibitem{lee2016training}
J.~H. Lee, T.~Delbruck, M.~Pfeiffer, Training deep spiking neural networks
  using backpropagation, Frontiers in Neuroscience 10 (2016) 508.

\bibitem{wu2018spatio}
Y.~Wu, L.~Deng, G.~Li, J.~Zhu, L.~Shi, Spatio-temporal backpropagation for
  training high-performance spiking neural networks, Frontiers in Neuroscience
  12 (2018) 331.

\bibitem{chakraborty2021fully}
B.~Chakraborty, X.~She, S.~Mukhopadhyay, A fully spiking hybrid neural network
  for energy-efficient object detection, IEEE Transactions on Image Processing
  30 (2021) 9014--9029.

\bibitem{kim2022beyond}
Y.~Kim, J.~Chough, P.~Panda, Beyond classification: directly training spiking
  neural networks for semantic segmentation, Neuromorphic Computing and
  Engineering 2~(4) (2022) 044015.

\bibitem{wu2021training}
H.~Wu, Y.~Zhang, W.~Weng, Y.~Zhang, Z.~Xiong, Z.-J. Zha, X.~Sun, F.~Wu,
  Training spiking neural networks with accumulated spiking flow, in:
  Proceedings of the AAAI Conference on Artificial Intelligence, 2021.

\bibitem{zhu2022event}
L.~Zhu, X.~Wang, Y.~Chang, J.~Li, T.~Huang, Y.~Tian, Event-based video
  reconstruction via potential-assisted spiking neural network, in: Proceedings
  of the IEEE/CVF Conference on Computer Vision and Pattern Recognition, 2022,
  pp. 3594--3604.

\bibitem{lee2020spike}
C.~Lee, A.~K. Kosta, A.~Z. Zhu, K.~Chaney, K.~Daniilidis, K.~Roy,
  Spike-flownet: event-based optical flow estimation with energy-efficient
  hybrid neural networks, in: European Conference on Computer Vision, Springer,
  2020, pp. 366--382.

\bibitem{hagenaars2021self}
J.~Hagenaars, F.~Paredes-Vall{\'e}s, G.~De~Croon, Self-supervised learning of
  event-based optical flow with spiking neural networks, Advances in Neural
  Information Processing Systems 34 (2021) 7167--7179.

\bibitem{ranccon2022stereospike}
U.~Ran{\c{c}}on, J.~Cuadrado-Anibarro, B.~R. Cottereau, T.~Masquelier,
  Stereospike: Depth learning with a spiking neural network, IEEE Access 10
  (2022) 127428--127439.

\bibitem{cuadrado2023optical}
J.~Cuadrado, U.~Ran{\c{c}}on, B.~R. Cottereau, F.~Barranco, T.~Masquelier,
  Optical flow estimation from event-based cameras and spiking neural networks,
  Frontiers in Neuroscience 17 (2023) 1160034.

\bibitem{patel2021spiking}
K.~Patel, E.~Hunsberger, S.~Batir, C.~Eliasmith, A spiking neural network for
  image segmentation, arXiv preprint arXiv:2106.08921 (2021).

\bibitem{kim2020spiking}
S.~Kim, S.~Park, B.~Na, S.~Yoon, Spiking-yolo: Spiking neural network for
  energy-efficient object detection, in: Proceedings of the AAAI Conference on
  Artificial Intelligence, Vol.~34, 2020, pp. 11270--11277.

\bibitem{horowitz20141}
M.~Horowitz, 1.1 computing's energy problem (and what we can do about it), in:
  2014 IEEE International Solid-State Circuits Conference Digest of Technical
  Papers, IEEE, 2014, pp. 10--14.

\bibitem{lemaire2022analytical}
E.~Lemaire, L.~Cordone, A.~Castagnetti, P.-E. Novac, J.~Courtois, B.~Miramond,
  An analytical estimation of spiking neural networks energy efficiency, in:
  International Conference on Neural Information Processing, Springer, 2022,
  pp. 574--587.

\bibitem{qu2023spiking}
J.~Qu, Z.~Gao, T.~Zhang, Y.~Lu, H.~Tang, H.~Qiao, Spiking neural network for
  ultralow-latency and high-accurate object detection, IEEE Transactions on
  Neural Networks and Learning Systems (2024).

\end{thebibliography}

\end{document}